\documentclass[journal,twoside,web]{ieeecolor}
\usepackage{generic}
\usepackage{cite}
\usepackage{amsmath,amssymb,amsfonts}
\usepackage{algorithmic}

\usepackage{graphicx}
\usepackage{algorithm}

\usepackage{adjustbox}
\usepackage{threeparttable}
\usepackage{booktabs}
\usepackage{multirow}
\usepackage{tabularx}
\usepackage{subfigure}

\usepackage{hyperref}
    
\hypersetup{hidelinks=true}
\usepackage{textcomp}
\def\BibTeX{{\rm B\kern-.05em{\sc i\kern-.025em b}\kern-.08em
    T\kern-.1667em\lower.7ex\hbox{E}\kern-.125emX}}
\begin{document}
\title{Unified Multi-modal Diagnostic Framework \\with Reconstruction Pre-training and Heterogeneity-combat Tuning}
\author{Yupei Zhang, Li Pan, Qiushi Yang, Tan Li, and Zhen Chen
\thanks{This work was supported in part by the InnoHK program. (Yupei Zhang and Li Pan contributed equally to this work) (Corresponding author: Zhen Chen, e-mail: zhen.chen@cair-cas.org.hk.)}
\thanks{Y. Zhang and L. Pan are with Department of Pathology, The University of Hong Kong, Hong Kong SAR.}
\thanks{Q. Yang is with Department of Electrical Engineering, City University of Hong Kong, Hong Kong SAR.} 
\thanks{T. Li is with Department of Computer Science, The Hang Seng University of Hong Kong, Hong Kong SAR.} 
\thanks{Z. Chen is with Centre for Artificial Intelligence and Robotics (CAIR), HKISI, Chinese Academy of Sciences, Hong Kong SAR.}
}
\maketitle

\begin{abstract}
Medical multi-modal pre-training has revealed promise in computer-aided diagnosis by leveraging large-scale unlabeled datasets. However, existing methods based on masked autoencoders mainly rely on data-level reconstruction tasks, but lack high-level semantic information. Furthermore, two significant heterogeneity challenges hinder the transfer of pre-trained knowledge to downstream tasks, \textit{i.e.}, the distribution heterogeneity between pre-training data and downstream data, and the modality heterogeneity within downstream data. To address these challenges, we propose a Unified Medical Multi-modal Diagnostic (UMD) framework with tailored pre-training and downstream tuning strategies. Specifically, to enhance the representation abilities of vision and language encoders, we propose the Multi-level Reconstruction Pre-training (MR-Pretrain) strategy, including a feature-level and data-level reconstruction, which guides models to capture the semantic information from masked inputs of different modalities. Moreover, to tackle two kinds of heterogeneities during the downstream tuning, we present the heterogeneity-combat downstream tuning strategy, which consists of a Task-oriented Distribution Calibration (TD-Calib) and a Gradient-guided Modality Coordination (GM-Coord). In particular, TD-Calib fine-tunes the pre-trained model regarding the distribution of downstream datasets, and GM-Coord adjusts the gradient weights according to the dynamic optimization status of different modalities. Extensive experiments on five public medical datasets demonstrate the effectiveness of our UMD framework, which remarkably outperforms existing approaches on three kinds of downstream tasks.
\end{abstract}


\section{Introduction}
\label{sec:introduction}
\IEEEPARstart{R}{ecently}, deep learning techniques have shown advantages in computer-aided diagnosis, mainly relying on the knowledge of expert-annotated medical datasets \cite{chen2022recent, xie2021survey}. However, collecting high-quality annotations for medical data is time-consuming and costly, making it difficult to construct large-scale medical datasets, thereby restricting the performance of current diagnostic algorithms that benefit from the expert annotations \cite{wang2021annotation}. Instead, a rational alternative is to exploit the knowledge of large amounts of unlabeled medical data effectively, which enhances the diagnostic performance and broadens the application scenarios of diagnostic algorithms. In particular, self-supervised pre-training \cite{huang2023self} provides a \textit{pretrain-finetune} paradigm that first performs pre-training on large-scale unlabeled data for superior representation learning, and then conducts the fine-tuning on a small amount of labeled data to adapt to downstream tasks. 

Different from medical imaging with uni-modality, multi-modal medical data (\textit{e.g.}, medical images and text descriptions) can improve diagnostic accuracy for various diseases by incorporating additional cross-modal knowledge independent of manual labeling \cite{hervella2021self}. Multi-modal pre-training \cite{chen2022m3ae} aims to encourage the model to capture semantic information in a self-supervised manner. By regularizing models to inter-modality and intra-modality, multi-modal pre-training works can be broadly categorized into two groups: contrastive learning-based \cite{radford2021CLIP, yang2022vision} and masked autoencoder-based methods \cite{singh2022flava, kwon2022maskedsignal, li2023scaling}. Contrastive learning-based methods train models to differentiate between similar and dissimilar pairs of data samples. Note that the paradigm of pushing or pulling samples in contrastive learning requires extremely large batch sizes and suffers from low efficiency, and the model performances are highly affected by the tricky selection of positive pairs and negative pairs \cite{wang2021solving}.
Meanwhile, masked autoencoder-based methods randomly remove a large proportion of the original data and encourage the model to reconstruct them \cite{wang2023fremae}.
This pre-training strategy efficiently achieves notable improvement over various downstream tasks, benefiting from more abundant supervision \cite{qing2023mar}. 
Although advancements have been achieved by masked autoencoder-based methods, they still have two major limitations worthy of improving.

\begin{figure}[!t]
\centering
\includegraphics[scale=.37]{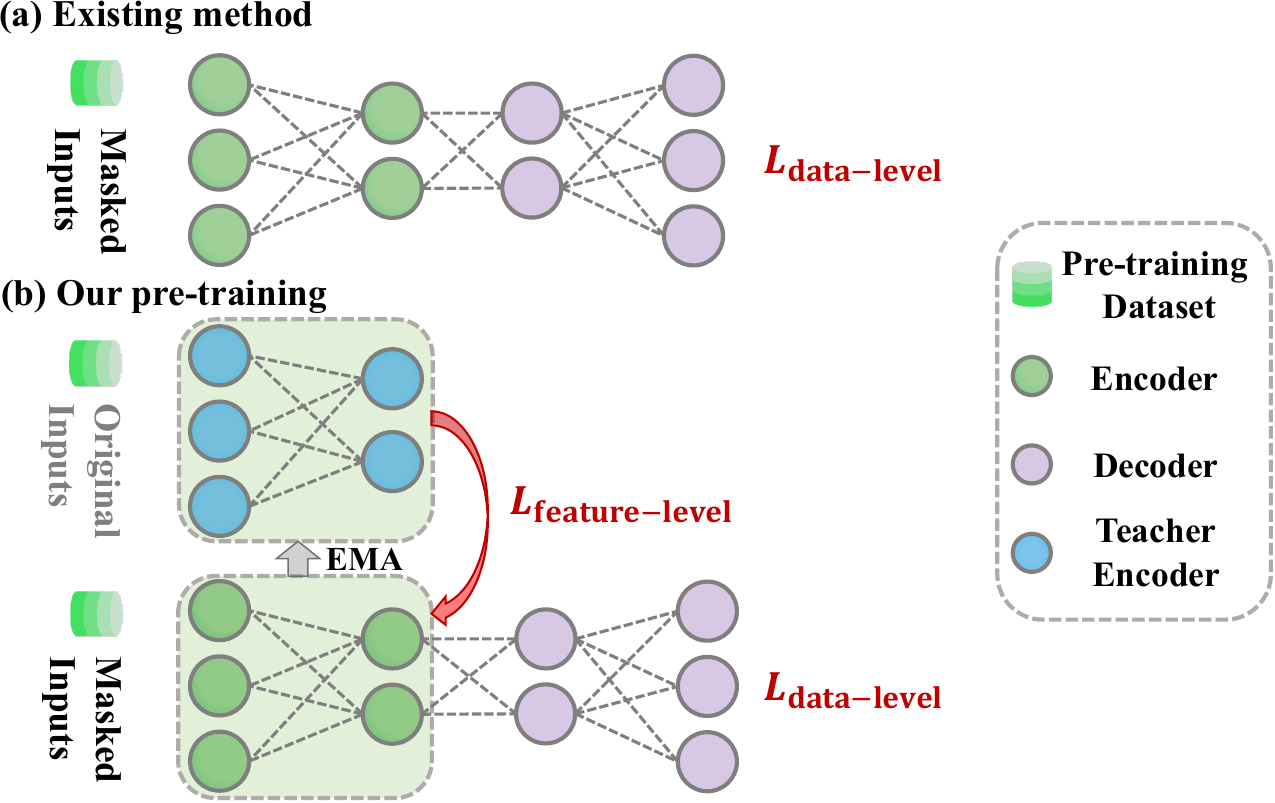} 
\caption{The comparison of pre-training strategies. Different from the existing methods (a) that aim for data-level reconstruction, we design a novel multi-level reconstruction pre-training (b) that enhances the encoder to learn transferable semantic features by incorporating data-level and feature-level reconstruction.} \label{fig:framediff1}
\end{figure}

The first limitation is the insufficient feature representation caused by using heuristic reconstruction targets, which may not fully capture the underlying structure of the data and result in insufficient pre-training. As illustrated in Fig. \ref{fig:framediff1} (a), most of the masked autoencoder-based methods \cite{he2022maskedMAE} simply employ original data (\textit{e.g.}, image pixels and text tokens) as prediction targets. However, this strategy can lead to overfitting of local statistics and high-frequency details that may be less relevant for data interpretation, such as background, illumination, and noise disturbance \cite{wei2022MaskFeat}.
To address the disadvantages of data-level reconstruction, recent studies attempt to use various manually extracted features as reconstruction targets. 
LocalMIM \cite{wang2023LocalMIM}, MaskFeat \cite{wei2022MaskFeat}, and FreMAE \cite{wang2023fremae} incorporate handcrafted feature descriptors (\textit{e.g.}, SIFT \cite{lowe1999objectSIFT}, HOG \cite{dalal2005histogramsHOG}, and Fourier spectrum \cite{wang2023fremae}) of the original data, to enhance the model's understanding on high-level features. Nevertheless, manually designed descriptors with specific strategies are sub-optimal and limit the model's generalization to other tasks and datasets. 
Rather than heuristically defining original data and handcrafted feature descriptors as reconstruction targets, we aim to further guide the model with masked inputs to reconstruct the high-level features of the original inputs. As shown in Fig. \ref{fig:framediff1} (b), two networks extract features of masked and original data respectively, and the target features of original data are dynamically adjusted by the model to perform the feature-level reconstruction, and accordingly, the difficulty of reconstruction task is modulated. By incorporating this feature-level reconstruction, our method enhances semantic understanding and improves feature representation learning.

\begin{figure}[!t]
\centering
\includegraphics[scale=.38]{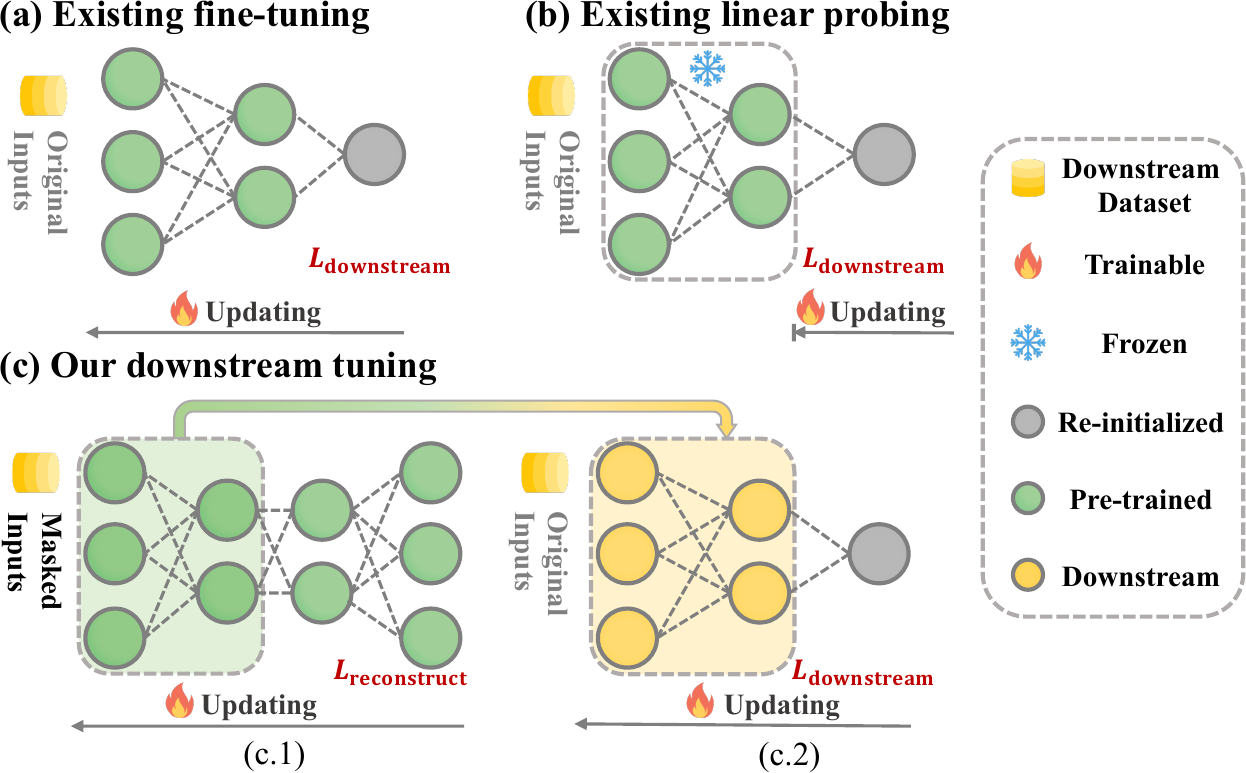} 
\caption{The comparison of fine-tuning strategies. Different from existing methods (a) and (b) that directly fine-tune the entire network or the final linear layer, we design a novel heterogeneity-combat downstream tuning (c) that promotes the encoder to learn semantic features of downstream data with reconstruction and boosts various downstream tasks.} \label{fig:framediff2}
\end{figure}

Another critical limitation is that existing pre-training works ignore the connection between the pre-training and fine-tuning stages, which hinders the knowledge transfer from pre-training to downstream tasks \cite{gururangan2020donPrefine}. 
Herein, we formulate this challenge from two perspectives, \textit{i.e.}, the distribution heterogeneity between pre-training and downstream data and the modality heterogeneity within multi-modal downstream data.
On the one hand, pre-training optimizes models to be robust on pre-training datasets, while the ultimate pursuit of model performance is for downstream task scenarios with data distribution shift \cite{gandelsman2022TTT, gururangan2020donPrefine}.  To perform the downstream diagnostic tasks, current medical pre-training methods \cite{kumar2022fine}, as shown in Fig. \ref{fig:framediff2} (a) and (b), directly fine-tune the entire network or the last linear layer of pre-trained model on the downstream datasets \cite{li2020MedVLPcomparison, chen2022m3ae}. 
However, simple fine-tuning may not be enough to bridge the distribution gap from pre-training to downstream datasets, resulting in even worse performance than specialized expert models on the target downstream task \cite{chang2022towards, nguyen2019overcomingMEVF}. 
On the other hand, during the fine-tuning phase, existing multi-modal pre-training methods \cite{su2023towards, kwon2022maskedsignal} jointly optimize the modules of different modalities. Nevertheless, since the rate of convergence varies for different modalities, this joint optimization strategy may lead the modules to the sub-optimum \cite{wang2020makesmultimodal},
where the potential of multi-modal data hasn't been fully exploited by multi-modal models \cite{peng2022balancedGM}. To address the above issues, we adapt the model for the specific downstream dataset through the consistent reconstruction task as in the pre-training stage, as shown in Fig. \ref{fig:framediff2} (c), to enhance the model's awareness toward the target distribution. Additionally, we present a dynamic gradient weighting mechanism for different modalities to ensure coordinated multi-modal training. By these means, our model can more effectively leverage pre-trained knowledge and adaptively balance the optimization of multiple modalities, ultimately improving performance on downstream medical diagnosis tasks.

In this work, we propose the Unified Medical Multi-modal Diagnostic (UMD) framework with Multi-level Reconstruction Pre-training (MR-Pretrain) and heterogeneity-combat downstream tuning strategies, which leverage the vast amounts of unlabelled medical data in pre-training, and bridge gaps in terms of distribution and modality. In addition to the data-level reconstruction supervision, we strengthen the constraints on the encoder by performing a novel feature-level reconstruction.
Specifically, we feed the original data into a teacher encoder and the masked data into a student encoder, where the student is supervised with the features extracted by the teacher in the feature space. This design enhances the encoder's representation learning for high-level semantic information and thus improves the generalization ability of pre-training. Moreover, to combat the inter-dataset distribution heterogeneity and inter-modality optimization interference, we propose a task-oriented distribution calibration (TD-Calib) module and a gradient-guided modality coordination (GM-Coord) module. TD-Calib calibrates the model trained on the pre-training datasets with instances from the downstream datasets in a mask-and-reconstruct manner, and GM-Coord dynamically adjusts the gradient weights of different modalities for coordinated multi-modal tuning. By enhancing the model's understanding of high-level semantic information with masked inputs and bridging the gap between pre-training and fine-tuning, the proposed UMD framework achieves superior performance on three kinds of downstream tasks. Our contributions are four-fold:
\begin{itemize}
\item To perform a more accurate diagnosis using medical multi-modal data, we introduce UMD, a novel unified framework that incorporates data-level and feature-level reconstruction to improve representation learning and heterogeneity-combat downstream tuning to bridge gaps in terms of distribution and modality.
\item To promote representation capabilities, we devise the MR-Pretrain strategy to enhance the multi-modal encoders with feature-level reconstruction in addition to data-level reconstruction. The MR-Pretrain feeds masked multi-modal samples into the student model to reconstruct target feature representations obtained from the teacher model with the original inputs, enabling the model to learn richer transferable multi-modal representations. 
\item To improve the transfer ability of the pre-trained model to various downstream tasks, we introduce the heterogeneity-combat downstream tuning, composed of two simple but effective modules, \textit{i.e.}, TD-Calib and GM-Coord. As such, our UMD framework can effectively bridge the distribution gap between pre-train data and downstream data, and thoroughly unleash the potential of multi-modal data.
\item We conduct experiments on three kinds of downstream tasks using five public multi-modal medical datasets. The results demonstrate the effectiveness of our UMD framework, which outperforms the state-of-the-art methods by a significant margin on all datasets.
\end{itemize}

\section{Related work}

\subsection{Multi-Modal Pre-Training}
Inspired by the great success achieved in uni-modal pre-training (\textit{e.g.}, natural language processing and computer vision), such as BERT \cite{devlin2018bert} and MAE \cite{he2022maskedMAE}, the multi-modal pre-training has gained increasing attention in recent years \cite{li2023blip}. The multi-modal pre-training aims to learn universal transferable representations from large-scale unannotated multi-modal data. Generally,
the inter-view and intra-view perspectives on the image and text lead to two main streams of pretext design for multi-modal pre-training, \textit{i.e.}, contrastive learning \cite{radford2021CLIP, yang2022vision} and masked multi-modal modeling \cite{singh2022flava, kwon2022maskedsignal, li2023scaling}. 

Contrastive learning trains models to maximize the similarity between positive pairs and minimize the similarity among negative pairs. Based on this simple idea, a large number of studies extend contrastive learning to perform self-supervised pre-training \cite{ radford2021CLIP}. Despite its effectiveness in learning useful representations, contrastive learning suffers from two drawbacks. Firstly, it demands a significant number of negative samples, which can be resource-intensive \cite{li2023scaling}. Secondly, it relies on the complex manual definition of positive and negative sample pairs \cite{wei2022MaskFeat}. 

Masked autoencoder is another type of paradigm for vision and language pre-training, which masks a portion of the input data and learns to recover the removed content \cite{he2022maskedMAE}. This mask-and-reconstruct strategy significantly reduces computational costs and encourages the model to learn data representations in a self-supervised manner \cite{devlin2018bert}.
Specifically, MAE \cite{he2022maskedMAE} demonstrated the self-supervised learning capability of masked autoencoders in computer vision by adopting the mask-and-reconstruct pretext which masked image patches. Singh \textit{et al.} \cite{singh2022flava} proposed unified pre-training schemes for the vision and language data by applying the mask-and-reconstruct to each modality. MaskVLM \cite{kwon2022maskedsignal} improved the existing vision-and-language pre-training approaches by alternately masking one modality to enhance the cross-modality alignment. DeepMIM \cite{ren2023deepmim} boosted the masked image modeling by the deep supervision of intermediate features to drive the shallower layers to learn meaningful representations.

As discussed above, most existing masked autoencoders set the original inputs as reconstruction targets. However, this data-level reconstruction strategy may cause overfitting to the low-level local statistics and high-frequency details, which can impede the model from capturing high-level semantic features from the inputs \cite{wei2022MaskFeat}.
To address this challenge, some recent studies attempt to improve feature-level supervision on the intermediate outputs of the encoders. For instance, MaskFeat \cite{wei2022MaskFeat} explicitly utilized the handcrafted image descriptors (\textit{e.g.}, HOG) as reconstruction targets to enhance feature representations. Wang \textit{et al.} \cite{wang2023LocalMIM} proposed a multi-scale reconstruction approach, which encourages the encoder to predict various handcrafted image descriptors at different layers. Yet, these methods bring strong assumptions on the reconstruction targets, which are biased and hard to be generalized due to the manual inputs.
Different from previous works, our UMD framework first performs pre-training with feature-level reconstruction to enhance feature representation learning, and then promotes fine-tuning stage through the tailored heterogeneity-combat downstream tuning.

\subsection{Medical Multi-Modal Pre-Training}
The medical multi-modal pre-training aims to improve the performance of diagnostic models by leveraging large-scale unlabeled multi-modal medical datasets \cite{chen2022m3ae}. On the one hand, unlike general computer vision datasets, medical datasets are naturally multi-modal, containing diverse imaging types and text data (\textit{e.g.}, diagnosis reports) \cite{pelka2018roco, subramanian2020medicat}. On the other hand, medical data requires manual annotations of human experts, which is time-consuming and costly \cite{wang2021annotation}. Therefore, there is a high demand for developing a self-supervised multi-modal pre-training method that can utilize unannotated medical data to improve the performance of existing deep learning models.

To achieve this goal, recent studies have explored self-supervised pre-training on medical datasets. For example, Li
\textit{et al.} \cite{li2020MedVLPcomparison} validated the effectiveness of medical multi-modal pre-training by evaluating four pre-trained vision-and-language models on medical datasets. To improve the performance of visual question-answering models, Khare \textit{et al.} \cite{khare2021mmbert} tokenized the medical images using convolutional neural networks to jointly pre-train both vision and language encoders under masked reconstruction modeling. Zhang \textit{et al.} \cite{zhang2022ConVIRT} utilized contrastive learning to pre-train models on paired medical images and texts, and evaluated pre-trained models on three medical imaging tasks, \textit{i.e.}, image classification, zero-shot image-image retrieval, and zero-shot text-image retrieval.  Endo \textit{et al.} \cite{endo2022gaitforemer} pre-trained a model on public datasets for gait movement forecasting, which can be further applied to clinical data to predict the severity of gait impairment for diagnosing Parkinson's disease.
Moon \textit{et al.} \cite{moon2022medvlpChest} presented a multi-modal attention masking approach to maximize generalization ability for both medical vision-language understanding tasks. Chen
\textit{et al.} \cite{chen2022m3ae} proposed a transformer-based pre-training model via multi-modal masked autoencoders, and achieved promising results on multiple multi-modal medical downstream tasks. 
On this basis, our UMD framework elaborately investigates the entire process from pre-training to fine-tuning, and leverages the characteristics of multi-modal medical data to facilitate the performance of medical diagnoses. 

\begin{figure*}[t!]
\centering
\includegraphics[scale=.27]{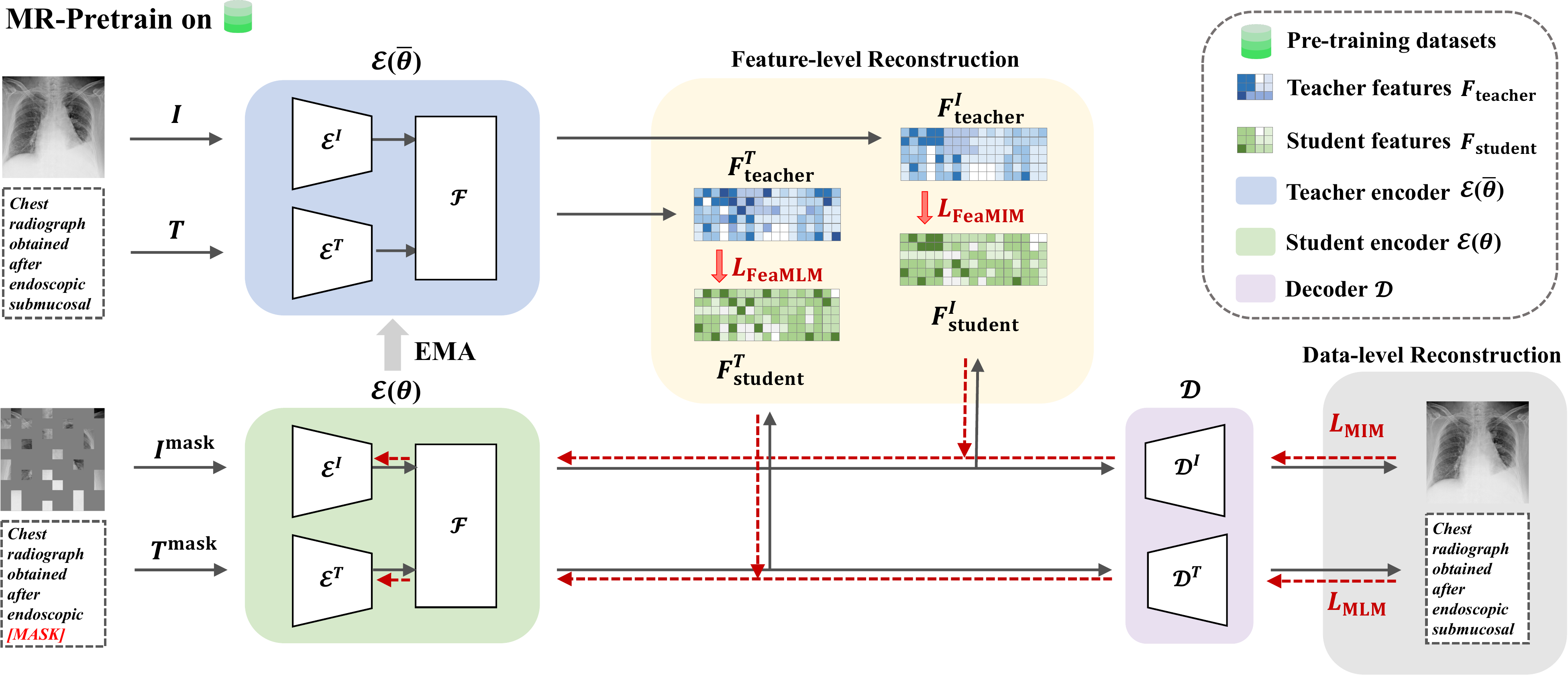}
\caption{Our MR-Pretrain exploits generalizable features from a large-scale unlabeled pre-training dataset in a dual-stream workflow. Besides the data-level reconstruction, we perform the feature-level reconstruction pretext task of features to encourage transferable representation learning.} \label{fig:pre-training}
\end{figure*}

\section{Methodology}

\subsection{Overview}
Our UMD framework contains two stages, \textit{i.e.}, MR-Pretrain in Fig.~\ref{fig:pre-training} to enhance the general feature representation and the heterogeneity-combat downstream tuning in Fig.~\ref{fig:fine-tuning} to boost the fine-tuning performance on various downstream tasks. In the MR-Pretrain stage, besides the widely-applied data-level reconstruction \cite{kwon2022maskedsignal}, we propose the feature-level reconstruction by a dual-stream workflow to encourage transferable representation learning from high-level features. In the heterogeneity-combat downstream tuning stage, {the TD-Calib bridges the distribution gap between the pre-training and downstream datasets, and the GM-Coord adjusts the gradient optimization of different modalities, thereby facilitating the performance of downstream tasks.}

\subsubsection{Model Architecture}
\noindent \textbf{Multi-modal encoder $\mathcal{E}$} comprises a student multi-modal encoder $\mathcal{E}(\theta)$ and a teacher multi-modal encoder $\mathcal{E}(\bar{\theta})$. They share the same network architecture, and each of them consists of a vision transformer (ViT)-based \cite{radford2021CLIP} vision encoder $\mathcal{E}^I$, a transformer-based \cite{liu2019roberta} language encoder $\mathcal{E}^T$, and a cross-attention-based multi-modal fusion module $\mathcal{F}$ \cite{chen2022m3ae}. For the multi-modal fusion module, we use two $N_m$-layer transformer models. Each model includes a self-attention layer for intra-modality learning, a cross-attention layer for inter-modality learning, and a feed-forward layer. The weights of teacher model $\mathcal{E}(\bar{\theta})$ is updated by the exponential moving average (EMA) \cite{tarvainen2017meanEMA} of the weights from student model $\mathcal{E}(\theta)$.

\noindent \textbf{Decoder $\mathcal{D}$} comprises a vision decoder $\mathcal{D}^I$ and a language decoder $\mathcal{D}^T$, and is designed to reconstruct the original image and text using the latent representations obtained through multi-modal fusion $\mathcal{F}$. Vision decoder $\mathcal{D}^I$ aims to reconstruct raw pixels that contain low-level textural information, while language decoder $\mathcal{D}^T$ is expected to recover the text tokens that represent high-level semantic information.
To this end, we employ a transformer \cite{vaswani2017attention} as the vision decoder for the low-level reconstruction, and the multi-layer perceptron (MLP) is utilized as the language decoder. 

\noindent \textbf{Downstream task head $\mathcal{H}$} for visual question-answering and image-text classification tasks, is a fully connected neural network with a layer normalization \cite{ba2016layer} and a GELU activation \cite{hendrycks2016gaussian}. 
For the image-text retrieval task, we utilize a linear layer as the head.

\subsubsection{Dataflow}
The input of our UMD framework consists of two streams: the image-text pair $(I, T)$ and the corresponding masked pair $(I^{\rm mask}, T^{\rm mask})$. The $I \in \mathbb{R}^{H\times W \times C}$ and $T \in \mathbb{R}^{L}$ represent the image and text, where $H$ and $W$ are image resolution, $C$ is the number of image channels, and $L$ is the length of a text sample. 
Following $\rm{M^3AE}$ \cite{chen2022m3ae}, we employ data sequentialization, linear projection embeddings, random masking, and position embeddings during the data preprocessing. A start-of-sequence token embedding and a special boundary token embedding are appended to the text sequence to indicate the beginning and end of the input text \cite{vaswani2017attention}. 
The vision $\mathcal{E}^I$ and language $\mathcal{E}^T$ encoder extract contextual representations of the image $H^I$ and text $H^T$ from the image-text pair $(I^{\rm mask}, T^{\rm mask})$. We further fuse $H^I$ and $H^T$ using the multi-modal fusion module $\mathcal{F}$, which produces multi-modal representations $Z^I = [z^I_{\rm CLS};z^I_1;z^I_2;...;z^I_{n^I}]$ for vision and $Z^T = [z^T_{\rm CLS};z^T_1;z^T_2;...;z^T_{n^T};z^T_{\rm SEP}]$ for language, where $n^I$ and $n^T$ are the numbers of image patches and text tokens, respectively. 
For data-level reconstruction of the MR-Pretrain stage and TD-Calib of the heterogeneity-combat downstream tuning stage, we feed the multi-modal representations $Z^I$ and $Z^T$ into $\mathcal{D}$, while for feature-level reconstruction, we feed them into linear layers. For GM-Coord, we feed the concatenated average embeddings of $Z^I$ and $Z^T$ into $\mathcal{H}$.


\subsection{Multi-Level Reconstruction Pre-Training} 

To promote the representation capabilities of the multi-modal encoders, we propose a multi-level reconstruction method in Fig.~\ref{fig:pre-training}, named MR-Pretrain, containing a novel feature-level reconstruction loss and a data-level reconstruction loss, encouraging the encoders to learn the high-level semantic representation from the unlabeled medical data.

\subsubsection{Data-Level Reconstruction}
The information density of language is poles apart from vision \cite{he2022maskedMAE, chen2022m3ae}, where the language is highly informative, and the vision is instead spatially redundant.
To eliminate the redundancies of vision and language while enabling the model to acquire valuable features from both vision and language in data-level reconstruction, we randomly mask image $I$ with a $75\%$ ratio and mask text $T$ with a $15\%$ masking ratio. 
Then we recover the masked inputs using the remaining data. The masked image modeling (MIM) loss and masked language modeling (MLM) loss are defined as follows:
\begin{equation} \label{eq:mim}
{L}_{\rm MIM} = \ \frac{1}{N} \sum_{n=1}^N (Y_{\rm MIM} - \ \mathcal{D}( \mathcal{E} \ (I^{\rm mask}_n, T^{\rm mask}_n ; \theta))^2,   
\end{equation}
\begin{equation} \label{eq:mlm}
 {L}_{\rm MLM} = - \frac{1}{N} \sum_{n=1}^N \log P_{\rm MLM}(Y_{\rm MLM} \ | \ I_n^{\rm mask}, T_n^{\rm mask}), 
\end{equation}
where $N$ is the total number of samples, $Y_{\rm MIM}$ is the raw pixel values of masked image patches, $Y_{\rm MLM}$ represents the labels of masked text tokens, and $P_{\rm MLM}$ represents the likelihood of each label given the input image-text pair.
By supervising the model to reconstruct details from masked image-text pairs, this data-level reconstruction task encourages the model to perceive low-level textual information without manual labels.

\subsubsection{Feature-Level Reconstruction}
Data-level reconstruction reduces the demand for data labeling, but entirely relying on this method may cause the model to overfit to the fine details of the input, hindering its ability to learn higher-level representations \cite{wang2023fremae, wang2023LocalMIM}. 
To address this problem, we propose the feature-level reconstruction that directly supervises the model in the feature space, encouraging the model to capture high-level semantic representations. Specifically, we employ a teacher model $\mathcal{E}(\bar{\theta})$ to extract the representations from original pairs $(I, T)$ and a student model $\mathcal{E}(\theta)$ for the masked image-text pairs $(I^{\rm mask}, T^{\rm mask})$. The outputs of the multi-modal encoder model contain two vectors, $Z^I$ for the image and $Z^T$ for the text. Accordingly, the outputs of the student model and teacher model are defined as follows:
\begin{equation} 
\begin{split}
    Z_n^I(\theta), Z_n^T(\theta) \ &= \ \mathcal{E} \ (I_n^{\rm mask}, T_n^{\rm mask} ; \theta), \\
    Z_n^I(\bar{\theta}), Z_n^T(\bar{\theta}) \ &= \ \mathcal{E} \ (I_n, T_n ; \bar{\theta}),
\end{split}
\end{equation}
where $n$ represents the index of the paired samples. Following \cite{chen2020simple}, we apply two linear layers $h^I$ and $h^T$ as projection heads to map the vision feature $Z_n^I$ and language feature $Z_n^T$ to a lower-dimensional latent space. The feature reconstruction loss for the image ${L}_{\rm FeaMIM}$ and text ${L}_{\rm FeaMLM}$ can be formulated as follows:
\begin{equation} \label{eq:fmim}
\begin{split}
{L}_{\rm FeaMIM} = \frac{1}{N} \sum^N_{n=1} ( h^I(Z_n^I(\bar{\theta})) - \ h^I(Z_n^I(\theta)))^2,  \\
{L}_{\rm FeaMLM} = \frac{1}{N} \sum^N_{n=1} (h^T(Z_n^T(\bar{\theta})) - \ h^T(Z_n^T(\theta)))^2,   
\end{split}
\end{equation}
where $N$ is the number of pre-training samples. Different from the data-level reconstruction \cite{chen2022m3ae, kwon2022maskedsignal} that encourages the model to concentrate on the low-level details, the proposed multi-modal feature reconstruction ${L}_{\rm FeaMIM}$ and ${L}_{\rm FeaMLM}$ guides the model to capture high-level semantic information from masked inputs. 
The teacher model weights $\bar{\theta}$ are updated under the exponential moving average \cite{tarvainen2017meanEMA} of the student model weights $\theta$, as follows:
\begin{equation} \label{e:EMA}
\bar{\theta}_t \ = \ \lambda \bar{\theta}_{t-1} + (1- \ \lambda)\theta_t,
\end{equation}
where $\lambda$ is the smoothing factor of teacher model $\bar{\theta}$ updating, and $t$ indicates the current iteration number. Given the full view of the input modalities, the teacher encoder $\mathcal{E}(\bar{\theta})$ can provide global feature-level guidance for the student encoder $\mathcal{E}(\theta)$, which encourages the student encoder $\mathcal{E}(\theta)$ to capture high-level semantic information during the pre-training.

\begin{figure*}[t!]
\centering
\includegraphics[scale=.27]{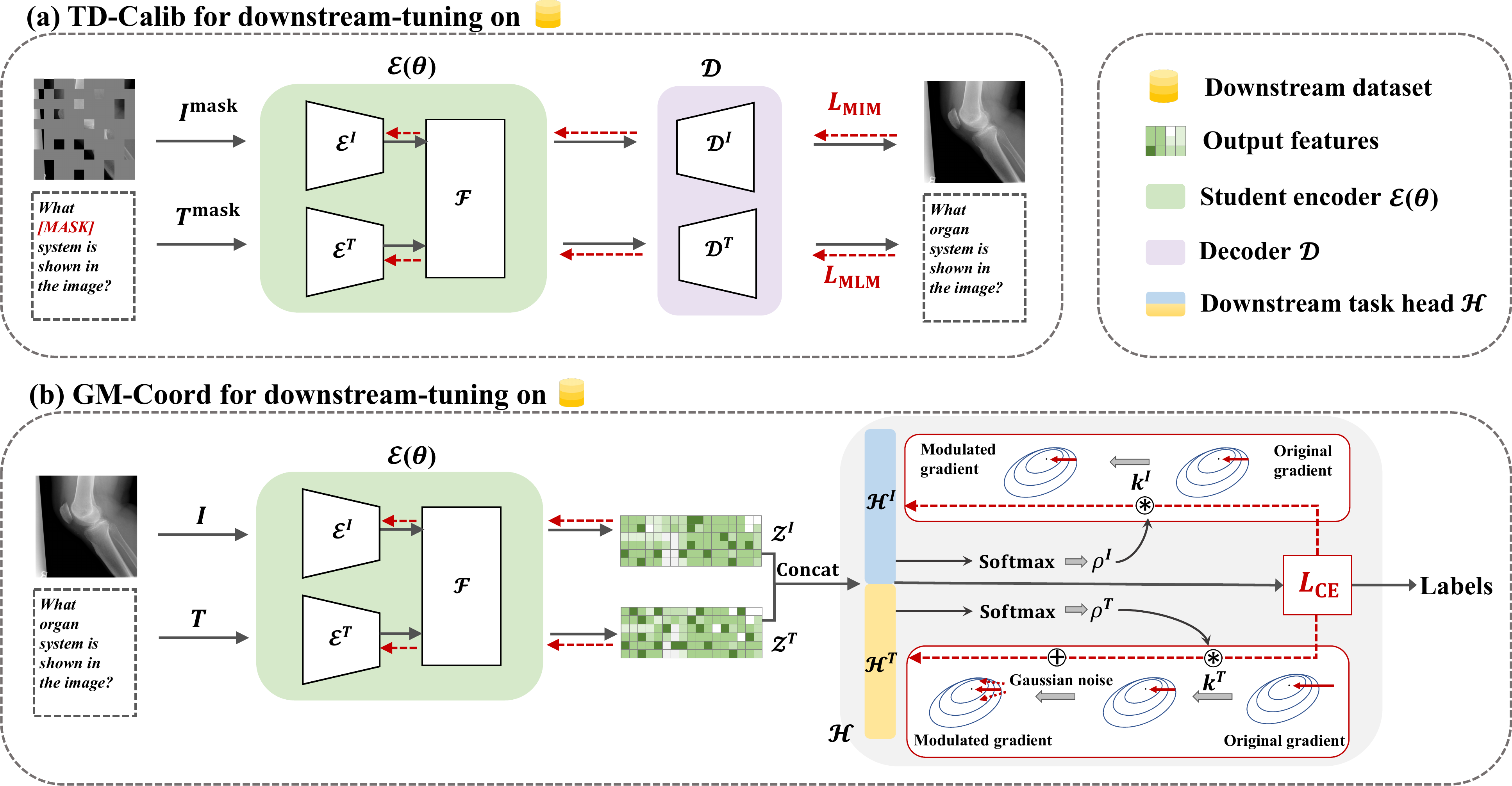}
\caption{Our heterogeneity-combat tuning facilitates medical diagnosis on downstream datasets. (a) The TD-Calib firstly calibrates the student multi-modal encoder to bridge the distribution gap, and then (b) the GM-Coord performs supervised fine-tuning to balance the modality optimization. For ease of understanding, we elaborate on the case of $\rho^T > 1$, where the gradient of the language modality should be modulated, as shown in (b).}
\label{fig:fine-tuning}
\end{figure*}

Furthermore, our MR-Pretrain incorporates the image-text matching (ITM) objective, a popular approach in vision-language understanding that aims to distinguish if a pair of image and text is matched. As such, our MR-Pretrain with ITM is effective for learning representations and improving the downstream performance \cite{liu2021unified}:
\begin{equation} \label{eq:itm}
 {L}_{\rm ITM} = - \frac{1}{N}\sum_{n=1}^N \log P_{\rm ITM}(Y_{\rm ITM} \ | \ I_n, T_n), 
\end{equation}
where $P_{\rm ITM}$ is the probability distribution obtained by applying a softmax function to the ITM decoder that consists of a linear layer, and $Y_{\rm ITM}$ represents the binary label for the ITM task. The value of one indicates a matched image-text pair, while zero indicates a mismatched pair. By promoting a joint representation of image and text inputs, our model enriches the correlation information between the modalities, thereby boosting the downstream tasks.

\subsubsection{MR-Pretrain Objective}
The total MR-Pretrain objective ${L}_{\rm pretrain}$ consists of five losses, \textit{i.e.}, ${L}_{\rm MIM}$, $ {L}_{\rm MLM}$, ${L}_{\rm FeaMIM}$, ${L}_{\rm FeaMLM}$, and ${L}_{\rm ITM}$, which is calculated as follows:
\begin{equation} \label{e:MR-Pretrainloss}
\begin{split}
    {L}_{\rm pretrain} \ = \ &(1 - \alpha) ({L}_{\rm MIM} \ + \ {L}_{\rm MLM}) \ + \\
    &\alpha ({L}_{\rm FeaMIM} \ + \ {L}_{\rm FeaMLM} \ ) \ + \ {L}_{\rm ITM},
\end{split}
\end{equation}
where $\alpha$ is a trade-off factor to balance the data-level and feature-level reconstruction.
By optimizing Eq.~\eqref{e:MR-Pretrainloss}, the model is pre-trained to learn image and text representation at multiple levels. In this way, our MR-Pretrain can improve the model's transferable representation for diverse downstream tasks, by leveraging the unlabeled pre-training data comprehensively.

\subsection{Heterogeneity-Combat Downstream Tuning}
The current \textit{pretrain-finetune} paradigm \cite{huang2023self, wei2022MaskFeat} directly fine-tunes the pre-trained model on the downstream datasets as illustrated in Fig.~\ref{fig:framediff2}. {However, this straightforward fine-tuning approach neglects the heterogeneity between pre-training and downstream datasets, as well as the modality heterogeneity within downstream optimization, resulting in sub-optimal performance on the specific dataset.} To tackle these two challenges, we propose the heterogeneity-combat downstream tuning in Fig.~\ref{fig:fine-tuning}, including TD-Calid to automatically calibrate the pre-trained encoder for a particular downstream dataset, and GM-Coord to dynamically balance the optimization of different modalities.

\subsubsection{Task-Oriented Distribution Calibration}
The distribution-heterogeneity inherently exists in pre-training and downstream data. A well-trained encoder carries abundant knowledge from the pre-training dataset, while how to efficiently transfer this pre-trained knowledge is an open-air question. 
Compared with the direct fine-tuning that is insufficient as the incoherent knowledge transfer from pre-training to downstream tuning, our TD-Calib module aims for a coherent transfer, which bridges the data distribution gap between the pre-training and the downstream tuning.
To this end, the pre-trained model is further trained on the downstream datasets by reconstructing the masked multi-modal data, as shown in Fig.~\ref{fig:fine-tuning} (a). This enables the model to adapt to the new distributions without explicit instruction of the ground truth, thus facilitating downstream objectives. Thus, we introduce the consistent pre-training objectives as TD-Calib training objectives, as follows:
\begin{equation} \label{e:TD-Calibloss}
\theta^*, \theta^*_1, ... \ , \theta^*_S \ = \ \mathop{\arg\min}\limits_{\theta, \theta_1, ... , \theta_S} \sum^S_{s=1} {L}_s(Y_s, \mathcal{D}_s(\mathcal{E}(I^{\rm mask}, T^{\rm mask} ; \theta_s))),
\end{equation}
where $Y_s$ represents the reconstruction targets of the masked image or text inputs, ${L}_s$ is the training objectives of the TD-Calib module, $S$ is the total number of training objectives that are empirically set as $4$, and s is the index of each training objective. In contrast to directly fine-tuning, our model optimized with Eq. \eqref{e:TD-Calibloss} not only adapts to the data distribution of the downstream domain, but also leverages the pre-trained knowledge to a greater extent.

\subsubsection{Gradient-Guided Modality Coordination}
Due to the modality heterogeneity, the optimization imbalance phenomenon exists in the joint training of multi-modal data, where the dominant modality suppresses the optimization of the other modalities during training. 
This phenomenon impacts the performance of medical multi-modal downstream tasks, such as visual question-answering and image-text classification tasks. 
To tackle this problem, we introduce the GM-Coord module to coordinate the optimization of each modality under the guidance of gradient changes, as illustrated in Fig.~\ref{fig:fine-tuning} (b).
During the GM-Coord, the contribution discrepancy among images and texts toward the learning objective is continuously monitored. The information is then utilized to adaptively modulate the gradients, thereby allocating more significant optimization updates to the suppressed modality.

For each multi-modal downstream task, the GM-Coord calculates the contribution $C^I$ of vision modality and $C^T$ of language modality. We split the task head $\mathcal{H} = [\mathcal{H}^I, \mathcal{H}^T]$ to separately measure the contribution of each modality, which is supervised by downstream task supervision. The calculation can be formulated with $u \in \{I, T\}$ for vision or language modality, as follows: 
\begin{equation} \label{e:encoderoutput}
    Z^I, Z^T \ = \ \mathcal{E}(I, T; \theta),
\end{equation}
\begin{equation} \label{e:contribution}
    C^u \ = \ {\rm softmax}(\mathcal{H}^u(Z^u))[j],
\end{equation}
where $C^u$ and $\mathcal{H}^u$ represent the contribution and task head for a specific modality respectively, and $[j]$ denotes a selection operator on the $j$-th class and j means the ground truth. In this way, $C^I$ means the prediction score on the ground truth class for the vision modality and $C^T$ means the prediction score on the ground truth class for the language modality. To quantify the optimization status of each modality, $\rho^I$ and $\rho^T$ are computed by the modality contribution, as follows: 
\begin{equation} \label{e:ratio}
    \rho^I \ = 
        \ \frac{\sum C^I}{\sum C^T}, \ \
        \rho^T \ = \ \frac{\sum C^T}{\sum C^I}.
\end{equation}
Then, we modulate the gradient of the modality that is optimizing fast. Taking the language modality as an example, the coordination coefficient $k^T$ is as follows:
\begin{equation} \label{e:coeffi}
    k^T \ = 
    \begin{cases}
        \ 1 \ - \ {\rm tanh}(\beta \cdot \rho^T), & \rho^T \ > \rho^I \\
        \ 1, & {\rm others}
    \end{cases}
\end{equation}
where the factor $\beta$ controls the degree of modulation and is set to 0.1. As such, the coordination coefficient $k^T<1$ when the optimization rate of one modality is higher than another, resulting in a reduction in the optimization speed of the faster modality. Finally, we integrate the coefficient $k^T$ into optimization together with Gaussian noise $\sigma(\theta) \sim \mathcal{N}(0, \Sigma^2_{\nabla \theta})$, where $\Sigma^2_{\nabla
\theta}$ represents the variance of the parameters' gradient, to improve the generalization ability. The modulated gradient $\tilde{g}$ of GM-Coord is as follows:
\begin{equation} \label{e:gm}
    \tilde{g}(\theta^T) \ \gets \ k^T \tilde{g}(\theta^T)
    + \ \sigma(\theta^T).
\end{equation}
Similarly, the gradients of vision modality $\tilde{g}(\theta^I)$ are also modulated following Eq.~\eqref{e:coeffi} and Eq.~\eqref{e:gm} if vision optimization is faster than the language optimization.
As such, the modulated gradients lead to balanced optimization of multi-modal data, thereby facilitating the performance of downstream tasks.

\begin{algorithm}[t] 
    \small{\caption{The pipeline of UMD} \label{pseudocode}} 
    \begin{algorithmic}[1]
        \small{\REQUIRE{Paired images and texts for Pre-training $\{I^P, T^P \}$ and for Downstream Tuning $\{I^D, T^D \}$; Model parameters $\Theta = \{\theta\} \cup \{\theta_s\}_{s=1}^{S}$; Random masking $M_I(\cdot)$ for images and $M_T(\cdot)$ for texts.}        
        \ENSURE{The trained optimal parameters $\Theta^*$}
        
        \COMMENT{MR-Pretrain}
        \WHILE{$\Theta$ doesn't reach convergence}
        \FOR{each $I$ and $T$ $\in$ $\{I^P, T^P \}$}
        \STATE $I^{\rm mask} \gets M_{I}(I)$; $T^{\rm mask} \gets M_{T}(T)$

        \STATE Minimize ${L}_{\rm pretrain}$ using Eq. \eqref{e:MR-Pretrainloss}
        \ENDFOR
        \ENDWHILE
        \STATE The $\mathcal{E} \circ \mathcal{D}$ is well-trained for MR-Pretrain stage.
        
        \COMMENT{TD-Calib}
        \WHILE{$\Theta$ doesn't reach convergence}
        \FOR{each $I$ and $T$ $\in$ $\{I^D, T^D \}$}
        \STATE $I^{\rm mask} \gets M_{I}(I)$; $T^{\rm mask} \gets M_{T}(T)$

        \STATE Minimize $\sum_{s=1}^{S} {L}_{s}(Y_s, \mathcal{D}_{s}(\mathcal{E}(I^{\rm mask}, T^{\rm mask} ; \theta))$ 
        \ENDFOR
        \ENDWHILE
        \STATE The $\mathcal{E} \circ \mathcal{D}$ is well-trained for TD-Calib module.
        
        \COMMENT{GM-Coord}
        \WHILE{$\Theta$ doesn't reach convergence} 
        \FOR{each $I$ and $T$ $\in$ $\{I^D, T^D \}$}
        \STATE Minimize ${L_{\rm {CE}}}(Y, \mathcal{H}(\mathcal{E}(I, T ; \theta))$

        \STATE Calculate $C^u$, $\rho^u$, $k^u$, $\tilde{g}(\theta^u)$ in Eq. \eqref{e:contribution}, \eqref{e:ratio}, \eqref{e:coeffi}, \eqref{e:gm} 

        \ENDFOR
        \ENDWHILE
        \STATE The $\mathcal{E} \circ \mathcal{H}$ is well-trained for GM-Coord module.}

    \end{algorithmic}
\end{algorithm}

\begin{table*}[t]
\parbox{.891\textwidth}{\caption{Comparison with state-of-the-art algorithms on three medical VQA datasets regarding accuracy metric. Best and second results are highlighted with \textbf{bold} and \underline{underline}.} \label{table:vqa}}
\centering
\begin{adjustbox}{center}
\begin{threeparttable}
\scalebox{0.9}{
\begin{tabular*}{.99\linewidth}{@{\extracolsep{\fill}}lccc cccc}
\toprule
\multirow{2}{*}{Methods} & \multicolumn{3}{c}{VQA-RAD} & \multicolumn{3}{c}{SLAKE} & VQA-Med-2019 \\
\cmidrule{2-8} 
  & Open & Closed & Overall & Open & Closed & Overall & Overall  \\
\cmidrule{1-8}
  MFB \cite{yu2017MFB} & 14.50 & 74.30 & 50.60 & 72.20 & 75.00 & 73.30 & - \\
  SAN \cite{yang2016SAN} & 31.30 & 69.50 & 54.30 & 74.00 & 79.10 & 76.00 & - \\
  BAN \cite{kim2018BAN} & 37.40 & 72.10 & 58.30 & 74.60 & 79.10 & 76.30 & - \\
\cmidrule{1-8}
  MEVF-BAN \cite{nguyen2019overcomingMEVF} & 49.20 & 77.20 & 66.10 & 77.80 & 79.80 & 78.60 & 77.86 \\
  CPRD-BAN \cite{liu2021CPRD} & 52.50 & 77.90 & 67.80 & 79.50 & 83.40 & 81.10 & - \\
\cmidrule{1-8}
  MAE \cite{he2022maskedMAE} & 67.04 & 77.94 & 73.61 & 77.21 & 82.21 & 79.17 & 73.60 \\
  CLIP \cite{radford2021CLIP} & 64.80 & 79.78 & 73.84 & 78.45 & 84.62 & 80.87 & 76.80 \\
  FLIP \cite{li2023scaling} & 65.92 & 78.31 & 73.39 & \underline{80.47} & 84.86 & 82.19 & 78.40 \\
  $\rm{M^3AE}$ \cite{chen2022m3ae} & \underline{67.23} & \underline{83.46} & \underline{77.01}  & 80.31 & \underline{87.82} & \underline{83.25} & \underline{79.87} \\
  UMD & \textbf{68.16} & \textbf{85.66} & \textbf{78.71} & \textbf{82.17} & \textbf{88.70} & \textbf{84.73} & \textbf{80.53} \\
\midrule
\bottomrule
\end{tabular*}
}
\end{threeparttable}
\end{adjustbox}
\end{table*}

\subsection{Algorithm Pipeline}
The training pipeline of our UMD framework is summarized in Algorithm \ref{pseudocode}, which includes the MR-Pretrain and heterogeneity-combat downstream tuning. We first perform the MR-Pretrain using Eq. \eqref{e:MR-Pretrainloss} on unannotated data, and obtain a pre-trained model that can generate general feature representations. Then, we conduct the TD-Calib in heterogeneity-combat downstream tuning using Eq. \eqref{e:TD-Calibloss}, which promotes the pre-trained model's smooth adaptation to downstream datasets. Finally, we perform the optimization of GM-Coord together with downstream objectives, enabling the model to capture semantic features of multi-modal data. The source code is available at \href{https://github.com/helenypzhang/UMD}{https://github.com/helenypzhang/UMD}.

\section{Experiment}
\subsection{Dataset}
We pre-train the model in our UMD framework using MedICaT \cite{subramanian2020medicat} and ROCO \cite{pelka2018roco} datasets and conduct the fine-tuning experiments on three downstream tasks, including three visual question-answering (VQA) datasets, one image-text retrieval dataset, and one image-text classification dataset.

\subsubsection{Pre-Training Datasets}
In our experimental setup, we conduct self-supervised pre-training on two datasets, \textit{i.e.}, MedICaT \cite{subramanian2020medicat} and ROCO \cite{pelka2018roco} dataset. 

\noindent \textbf{MedICaT dataset} \cite{subramanian2020medicat} comprises more than 217,000 medical images and their corresponding captions and inline textual references. Following $\rm{M^3AE}$ \cite{chen2022m3ae}, we randomly allocate 1,000 samples for test, 1,000 for validation, and the remaining data for training purposes.

\noindent \textbf{ROCO dataset} \cite{pelka2018roco} contains more than 81,000 medical radiology images, encompassing a variety of imaging modalities such as Computed Tomography (CT), X-ray, ultrasound, fluoroscopy, angiography, mammography, positron emission tomography, and Magnetic Resonance Imaging (MRI). Each image is accompanied by a corresponding caption. We follow the dataset splits in ROCO \cite{pelka2018roco}, with over 65,000 radiology images to the training set, over 8,000 radiology images to the validation set, and over 8,000 radiology images to the test set.

\subsubsection{Downstream Tuning Datasets}
We evaluate the effectiveness of our pre-training approach by conducting experiments on the VQA, image-text retrieval tasks, and image-text classification, utilizing the official split of each dataset in downstream experiments.
For the VQA task, we select three public datasets, \textit{i.e.}, VQA-RAD \cite{lau2018vqarad}, SLAKE \cite{liu2021slake}, and VQA-Med-2019 \cite{abacha2019vqamed2019}. 
Moreover, the ROCO dataset \cite{pelka2018roco} and MELINDA dataset \cite{wu2021melinda} are utilized in image-text retrieval and image-text classification tasks, respectively.

\noindent \textbf{VQA-RAD dataset} \cite{lau2018vqarad} includes 315 images, consisting of 104 axial single-slice CTs or MRIs for head, 107 X-rays for chest, and 104 axial CTs for abdomen, each accompanied by corresponding captions. There are over 3.5K visual questions in VQA-RAD, including open-ended and closed-ended answer types. In particular, there are 3,064 question-answer pairs in training set, 451 question-answer pairs in validation set, and 451 question-answer pairs in test set.

\noindent \textbf{SLAKE dataset} \cite{liu2021slake} comprises 642 multi-modal images covering 12 diseases and 39 organs of the human body to ensure dataset diversity. The question-answer pairs are 14K. Both open-ended and closed-ended answer types are included in the SLAKE and VQA-RAD datasets, determined by whether the answer choices are limited or not. In particular, the dataset is divided into training, validation, and test sets with the ratio of $75\%$, $15\%$ and $15\%$.

\noindent \textbf{VQA-Med-2019 dataset} \cite{abacha2019vqamed2019} is composed of 4,200 radiological images and 15,292 question-answer pairs.
The dataset is split into training set with 3,200 images, validation set with 500 images, and test set with 500 images.

\noindent \textbf{ROCO dataset} \cite{pelka2018roco} is utilized on the image-text retrieval task. The image-text retrieval task comprises two subtasks: image-to-text retrieval and text-to-image retrieval. The former aims to retrieve the most relevant texts based on the given image, while the latter aims to retrieve the most relevant images based on the given text. 

\noindent \textbf{MELINDA dataset} \cite{wu2021melinda}, which contains 2,833 figures paired with corresponding detailed sub-figures and sub-captions, is utilized on the image-text classification task. The dataset is split into train, validation, and test sets, with the ratio of $80\%$, $10\%$ and $10\%$.

\begin{table*}[t]
\parbox{.81\textwidth}{\caption{Comparison with state-of-the-art algorithms on medical image-text retrieval task on ROCO dataset. Best and second results are highlighted with \textbf{bold} and \underline{underline}.} \label{table:irtr}}
\centering
\begin{adjustbox}{center}
\begin{threeparttable}
\scalebox{0.92}{
\begin{tabular*}{.90\linewidth}{@{\extracolsep{\fill}} l  ccc  ccc}
\toprule
\multirow{2}{*}{Methods} & \multicolumn{3}{c}{Text-to-image retrieval} & \multicolumn{3}{c}{Image-to-text retrieval} \\
\cmidrule{2-7} 
& R@1 & R@5 & R@10 & R@1 & R@5 & R@10 \\
\cmidrule{1-7}
ViT+BERT \cite{dou2022VitbertMeter} & 5.25 & 15.85 & 25.85 & 6.85 & 21.25 & 31.60 \\
ViLT \cite{kim2021vilt} & 9.75 & 28.95 & 41.40 & 11.90 & 31.90 & 43.20 \\
METER \cite{dou2022VitbertMeter} & 11.30 & 27.25 & 39.60 & 14.45 & 33.30 & 45.10 \\
\cmidrule{1-7}
MAE \cite{he2022maskedMAE} & 4.35 & 17.96 & 28.96 & 4.95 & 18.31 & 28.06 \\
CLIP \cite{radford2021CLIP} & 14.41 & 39.67 & 54.68 & 17.61 & 42.92 & 57.98 \\
FLIP \cite{li2023scaling} & 17.66 & 46.62 & 61.03 & 17.46 & 45.57 & 61.53 \\
$\rm{M^3AE}$ \cite{chen2022m3ae} & \underline{22.20} & \underline{52.50} & \underline{66.65} & \underline{22.90} & \underline{51.05} & \underline{65.80} \\
UMD & \textbf{23.21} & \textbf{54.28} & \textbf{67.88} & \textbf{24.39} & \textbf{54.27} & \textbf{68.97} \\
\midrule
\bottomrule
\end{tabular*}
}
\end{threeparttable}
\end{adjustbox}
\end{table*}

\subsection{Implementation Details}
Our experiments are implemented with the PyTorch Lightning library \cite{falcon2019pytorch} on three NVIDIA A100 PCIe 40 GB GPUs. Details of each task are elaborated as follows.
\subsubsection{Multi-Level Reconstruction Pre-Training} 
For MR-Pretrain, we train $\mathcal{E} \circ \mathcal{D}$ end-to-end. We start from the CLIP-ViT-B model \cite{radford2021CLIP} as the vision encoder, the RoBERTa-base \cite{liu2019roberta} as the language encoder, with the multi-modal fusion module provided by $\rm{M^3AE}$ \cite{chen2022m3ae}. The multi-modal module consists of 6 Transformer layers with a hidden state dimension of 768 and 12 heads. We use AdamW optimizer \cite{loshchilov2017adamw} to train the models for 100,000 steps, with a learning rate of $1\times10^{-5}$ for the uni-modal encoders and $5\times10^{-5}$ for the multi-modal fusion module. We set the warm-up ratio to $10\%$, with a linear learning rate scheduler after warm-up. To resize each image, we use a center-crop method with a size of $314\times 314$. The trade-off factor $\alpha$ in MR-Pretain is set as $0.5$. The smoothing factor $\lambda$ of EMA is 0.995 for weight updating.

\subsubsection{Heterogeneity-combat Downstream Tuning}
For TD-Calib, we fine-tune the $\mathcal{E} \circ \mathcal{D}$ end-to-end. In order to bridge the data distribution gap between pre-training and fine-tuning, we conduct TD-Calib-guided downstream tuning. Specifically, we initialize the multi-modal encoder with the pre-trained weights, and feed images and texts to the model to further pre-train both $\mathcal{E}$ and $\mathcal{D}$. The masking ratio is set to $75\%$ for images and $15\%$ for texts. Moreover, the warm-up ratio is $10\%$, with a linear learning rate scheduler used after warm-up steps. We use AdamW as the optimizer with a weight decay of 0.01 for all downstream tasks. The initial learning rate for VQA-RAD, SLAKE, VQA-Med-2019, ROCO, and MELINDA is set to $1\times10^{-5}$, $5\times10^{-6}$, $5\times10^{-6}$, $1\times10^{-5}$ and $1\times10^{-5}$, respectively, and linearly decay to zero during training.

For GM-Coord, we fine-tune $\mathcal{E} \circ \mathcal{H}$ end-to-end.
Specifically, we further fine-tune the multi-modal encoder $\mathcal{E}$ optimized by the TD-Calib module under different downstream tasks, together with the downstream task-specific head $\mathcal{H}$. For each downstream task, we guarantee the fairness of the experiment by adopting the same $\mathcal{H}$ for different comparison methods.
We utilize the AdamW optimizer with an initial learning rate of $5 \times 10^{-6}$, a warm-up ratio of $10\%$, and a linear decay during training for VQA-Med-2019, while for other downstream datasets, we use cosine decay. The weight decay is set to $0.01$ for SLAKE and MELINDA datasets, and $0.1$ for VQA-RAD, VQA-Med-2019 and ROCO datasets.

\subsubsection{Evaluation Metric}
To conduct a comprehensive evaluation, we analyze diverse performance metrics on different downstream tasks.
We follow the previous study \cite{chen2022m3ae} to adopt the accuracy for the VQA and image-text classification tasks, and Recall@K with K=1, 5 and 10 for the image-text retrieval task, respectively. In VQA and image-text classification, the \textit{Overall} term specifically refers to the micro-average accuracy of both open-ended and closed-ended questions. In addition, the Recall@K, commonly used in information retrieval tasks, is an evaluation metric that measures the proportion of relevant items that are retrieved in the top K results. In other words, the Recall@K measures how many of the relevant items are actually retrieved in the top K results. We conduct experiments for Recall@K with K=1, 5 and 10, which represent the proportion of relevant items retrieved in different predictions.

\subsection{Downstream Experiments}

\begin{table}[t]
\centering
\parbox{.99\linewidth}{\caption{Comparison with state-of-the-art algorithms on medical image-text classification task on MELINDA dataset. Best and second results are highlighted with \textbf{bold} and \underline{underline}.} \label{tab:cls}}
\resizebox{.60\linewidth}{!}{ 
\begin{tabular}{llc}
\toprule
Modalities & Methods & Accuracy\\
\midrule
Image-only & ResNet-101 \cite{he2016RESNET101} & 63.84 \\ 
\cmidrule{1-3}
\multirow{3}{*}{Text-only} & LSTM \cite{hochreiter1997LSTM} & 59.20 \\
& RoBERTa \cite{liu2019roberta} & 75.40 \\
& SciBERT \cite{beltagy2019scibert} & 77.70 \\
\cmidrule{1-3}
\multirow{4}{*}{Multi-modal} & NLF \cite{wu2021melinda} & 76.60 \\ 
& SAN \cite{yang2016SAN} & 72.30 \\
& ViLBERT \cite{lu2019vilbert} & \underline{78.60} \\
& MAE \cite{he2022maskedMAE} & 78.03 \\
& CLIP \cite{radford2021CLIP} & 77.16 \\
& FLIP \cite{li2023scaling} & 77.36 \\
& $\rm{M^3AE}$ \cite{chen2022m3ae} & 78.50 \\
& UMD & \textbf{79.58} \\ 
\midrule
\bottomrule
\end{tabular}}
\end{table}

\subsubsection{Medical Visual Question-Answering}
The VQA is a multi-modal task that requires both images and questions as input, and is expected to answer questions about medical images. The VQA questions can belong to either open or closed categories, where open-category questions require the model to generate a free-form answer, while closed-category questions require the model to select a predefined answer from a set of options. Medical VQA task is particularly useful in medical diagnosis and treatment planning, where doctors often rely on visual information to make informed decisions. 
As shown in Table \ref{table:vqa}, our UMD framework outperforms state-of-the-art models on all VQA datasets, achieving the accuracy of 68.16\%, 85.66\%, and 78.71\% for VQA-RAD, 82.17\%, 88.70\%, and 84.73\% for SLAKE, and 80.53\% for VQA-Med-2019. 
In terms of \textit{Overall} performance, our UMD framework outperforms the second-best method (marked in underline)  by 1.70\%, 1.48\%, and 0.66\% on three VQA datasets. 
These improvements contribute to the transferable features learned by our tailored MR-Pretrain and the heterogeneity-combat downstream tuning stages. 

\begin{figure}[t]
\centering
\includegraphics[scale=.199]{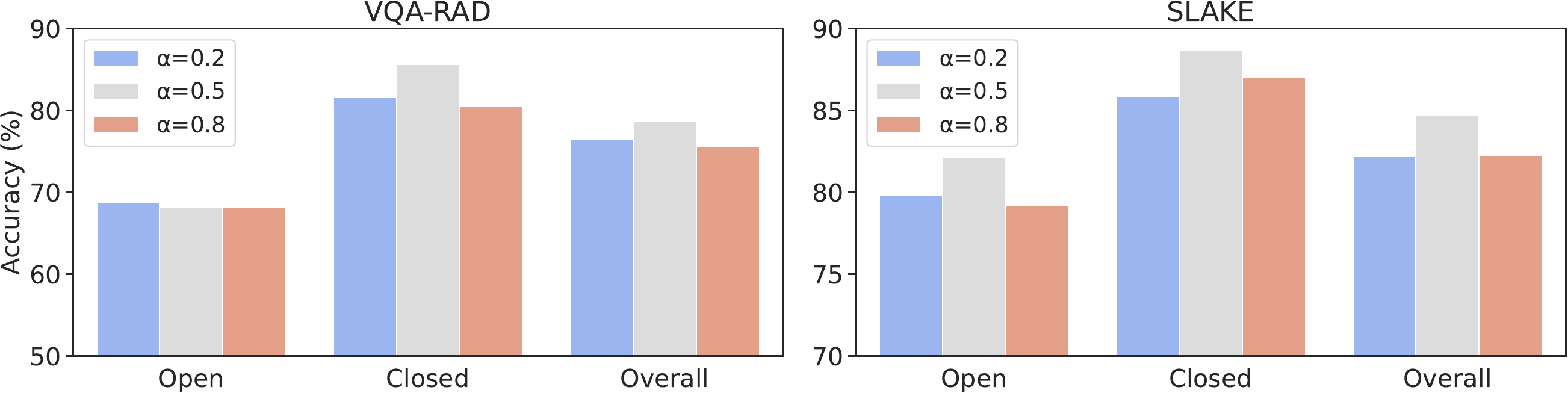}
\caption{Ablation study on the hyper-parameter $\alpha$ in MR-Pretrain. Our UMD framework achieves the best performance when $\alpha$ is set as $0.5$.} \label{fig:alpha}
\end{figure}

\begin{figure*}[t!]
\centering
\includegraphics[scale=.25]{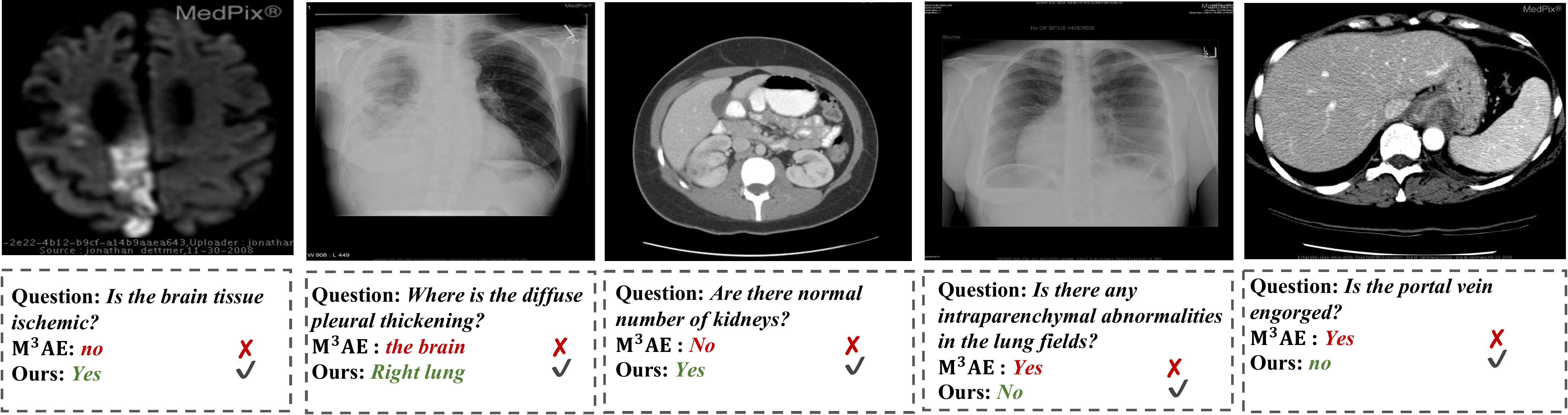}
\caption{Visualization of medical VQA comparison on VQA-RAD dataset. Our UMD framework is capable of providing more accurate answers to medical questions of different difficulties.} \label{fig:vqa2019test}
\end{figure*}

\begin{table*}[htbp]
\caption{Ablation study of UMD on three medical VQA datasets. \label{table:abvqa}}
\centering
\begin{adjustbox}{center}
\begin{threeparttable}
\scalebox{0.90}{
\begin{tabular*}{.95\linewidth}{@{\extracolsep{\fill}}l ccc |ccc |ccc| c}

\toprule
& Pre-training & \multicolumn{2}{c|}{Fine-tuning} & \multicolumn{3}{c|}{VQA-RAD} & \multicolumn{3}{c|}{SLAKE} & VQA-Med-2019 \\
\cmidrule{2-11}
& MR-Pretrain & TD-Calib & GM-Coord & Open & Close & \textbf{Overall} & Open & Close & \textbf{Overall} & \textbf{Overall} \\
\midrule
1 & & & & 65.36 & 78.68 & 73.39 & 74.88 & 78.13 & 76.15 & 72.00 \\				
2 & \checkmark & & & \textbf{69.27} & 82.72 & 77.38 & 81.86 & 86.06 & 83.51 & 77.07 \\
3 & & \checkmark & & 66.48 & 80.51 & 74.95 & 79.84 & 85.10 & 81.90 & 74.67 \\
4 & & & \checkmark & 68.16 & 80.51 & 75.61 & 77.36 & 86.54 & 80.96 & 76.00 \\
5 & & \checkmark & \checkmark & 67.60 & 81.99 & 76.27 & 80.00 & 86.54 & 82.56 & 77.87 \\
6 & \checkmark & \checkmark &  & 68.16 & 84.19 & 77.83 & 81.40 & 87.74 & 83.88 & 79.47\\
7 & \checkmark &  & \checkmark & \textbf{69.27} & 83.82 & 78.05 & \textbf{82.95} & 86.54 & 84.35 & 80.23 \\
8 & \checkmark & \checkmark & \checkmark & 68.16 & \textbf{85.66} & \textbf{78.71} & 82.17 & \textbf{88.70} & \textbf{84.73} & \textbf{80.53}\\
\midrule
\bottomrule

\end{tabular*} 
}
\end{threeparttable}
\end{adjustbox}
\end{table*}

We also conduct experiments using strong baselines of MAE, CLIP, and FLIP algorithms. The models are first pre-trained with different training objectives on the MedICaT and the ROCO datasets, and then fine-tuned with the same prediction head with the cross-entropy loss on VQA datasets. We adopt the same backbones and task heads to ensure fairness in the experiment.
Compared with four strong baselines, two of which are masked autoencoder-based (\textit{i.e.}, MAE and $\rm{M^3AE}$), and the other two are contrastive learning-based pre-training methods (\textit{i.e.}, CLIP and FLIP), our accuracy increases by 6.93\%, 0.66\%, 3.73\%, and 2.13\%, on the VQA-Med-2019 dataset, respectively. These results show that UMD is not only superior to masked autoencoder-based pre-training, but also outperforms other types of pre-training algorithms.

\noindent \textbf{Hyper-parameters Analysis.}
We further conduct experiments on one of the most significant hyper-parameters, \textit{i.e.}, $\alpha$ in Eq. \eqref{e:MR-Pretrainloss}, to investigate the trade-off between feature-level and data-level reconstruction in MR-Pretrain and TD-Calib. In our hyper-parameters study, we set $\alpha$ as $0.2$, $0.5$, and $0.8$ both for VQA-RAD and SLAKE datasets. As illustrated in Fig. \ref{fig:alpha}, our UMD framework achieves the best performance when $\alpha$ is set as 0.5, further demonstrating the rationality of the hyper-parameter setting in our UMD framework.

\subsubsection{Medical Image-Text Retrieval} 
Image-text retrieval is a cross-modal task including medical image-to-text retrieval and medical text-to-image retrieval tasks, requiring the model to exploit useful information across modalities. The experimental results are presented in Table \ref{table:irtr}. We perform comprehensive comparisons with state-of-the-art methods, including ViLT, METER, MAE, CLIP, FLIP, and $\rm{M^3AE}$. The results show that our UMD framework achieves the best R@K (K=1,5 and 10) performances of 23.21\%, 54.28\% and 67.88\% for text-to-image retrieval, and 24.39\%, 54.27\%, and 68.97\% for image-to-text retrieval task. UMD surpasses the second-best one by 1.78\% and 3.22\% in terms of R@5 text-to-image and image-to-text retrieval tasks respectively. These experimental results show the effectiveness of our UMD framework on medical image-text retrieval.

\subsubsection{Medical Image-Text Classification} 
Image-text classification aims to give a label to a medical image-text pair, which also belongs to the multi-modal task. By training a model that can classify medical images associated with the text descriptions, this task is beneficial in medical research and clinical scenarios. Besides the baselines with image-only data and text-only data, we perform the comparison with advanced multi-modal methods ViLBERT, CLIP, and FLIP in the general domain, and $\rm{M^3AE}$ in the medical multi-modal domain. As shown in Table \ref{tab:cls}, our UMD framework achieves the best accuracy of $79.58\%$ on the MELINDA dataset, outperforming the second-best {ViLBERT} by $0.98\%$. The remarkable advantage in image-text classification demonstrates the effectiveness of our UMD framework on medical multi-modal data.

\begin{table}[tbp]
\caption{Ablation study of UMD on MELINDA dataset.\label{table:abcls}}
\centering
\begin{adjustbox}{center}
\begin{threeparttable}
\scalebox{0.78}{
\begin{tabular}{@{\extracolsep{\fill}}l ccc | c ccc}
\toprule
& Pre-training & \multicolumn{2}{c|}{Fine-tuning} & \multirow{2}{*}{Accuracy} & \multirow{2}{*}{AUC} & \multirow{2}{*}{Sensitivity} & \multirow{2}{*}{Specificity} \\
\cmidrule{2-4}
& MR-Pretrain & TD-Calib & GM-Coord & \\
\midrule
1 & & & & 66.09 & 78.20 & 27.16 & 90.86 \\
2 & \checkmark & & & 78.55 & 85.48 & 34.59 & 94.16 \\
3 & & \checkmark & & 74.91 & 86.98 & 32.23 & 92.94 \\
4 & & & \checkmark & 69.72 & 81.74 & 31.11 & 91.70 \\
5 & & \checkmark & \checkmark & 75.26 & 84.28 & 33.83 & 93.49 \\
6 & \checkmark & \checkmark &  & 78.72 & 89.13 & 34.16 & 94.14 \\
7 & \checkmark &  & \checkmark & 79.24 & 89.39 & 34.59 & 94.27 \\
8 & \checkmark & \checkmark & \checkmark & \textbf{79.58} & \textbf{91.52} & \textbf{36.00} & \textbf{94.56} \\
\midrule
\bottomrule

\end{tabular} 
}
\end{threeparttable}
\end{adjustbox}
\end{table}

\subsection{Ablation Study}

To quantitatively evaluate the effectiveness of our proposed components, \textit{i.e.}, MR-Pretrain, TD-Calib and GM-Coord, we conduct ablation studies for each component on three VQA datasets and one image-text classification dataset. The ablation results are illustrated in Table \ref{table:abvqa} and Table \ref{table:abcls}.

\begin{itemize}
\item Line 1: The baseline simply trains the multi-modal encoder and downstream task head from scratch, without relying on pre-trained models or the proposed components. This baseline serves as a performance lower bound for VQA and medical image-text classification.

\item Line 2-4: The proposed components (\textit{i.e.}, MR-Pretrain, TD-Calib without pre-training, and GM-Coord without pre-training) are individually added on the basis of the baseline (Line 1). These records can validate the independent impact of these three components.
\item Line 5-8: The possible combination of the proposed three components. These records are crucial for the ablation study of MR-Pretrain, TD-Calib and GM-Coord.
\end{itemize}

\begin{figure*}[t!]
\centering
\subfigure[2D Clustering of FLIP]{
\label{2d.3}
\includegraphics[scale=.33]{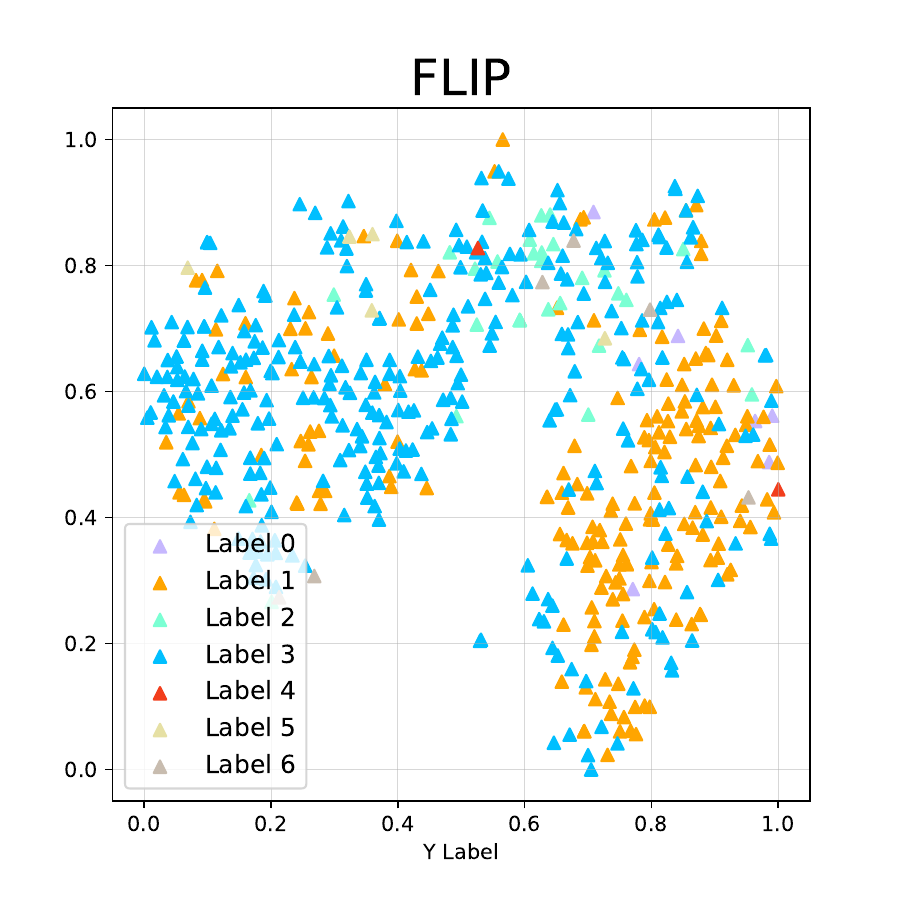}}
\subfigure[2D Clustering of $\rm{M^3AE}$]{
\label{2d.1}
\includegraphics[scale=.33]{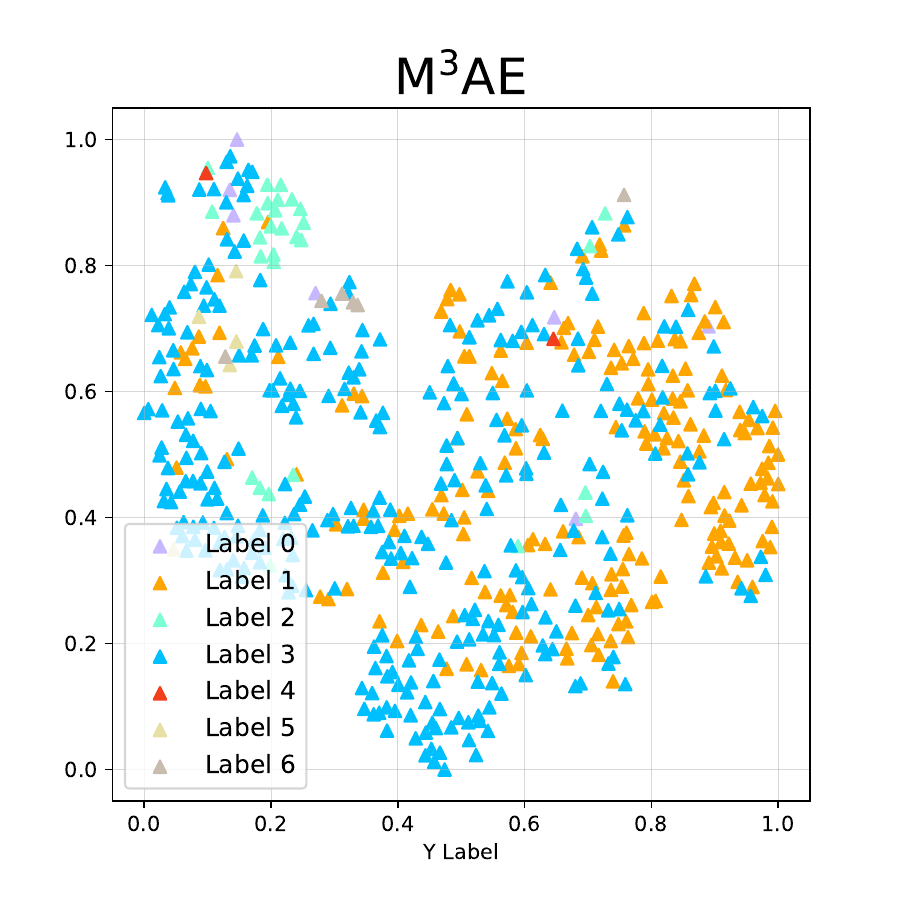}}
\subfigure[2D Clustering of Ours (UMD)]{
\label{2d.2}
\includegraphics[scale=.33]{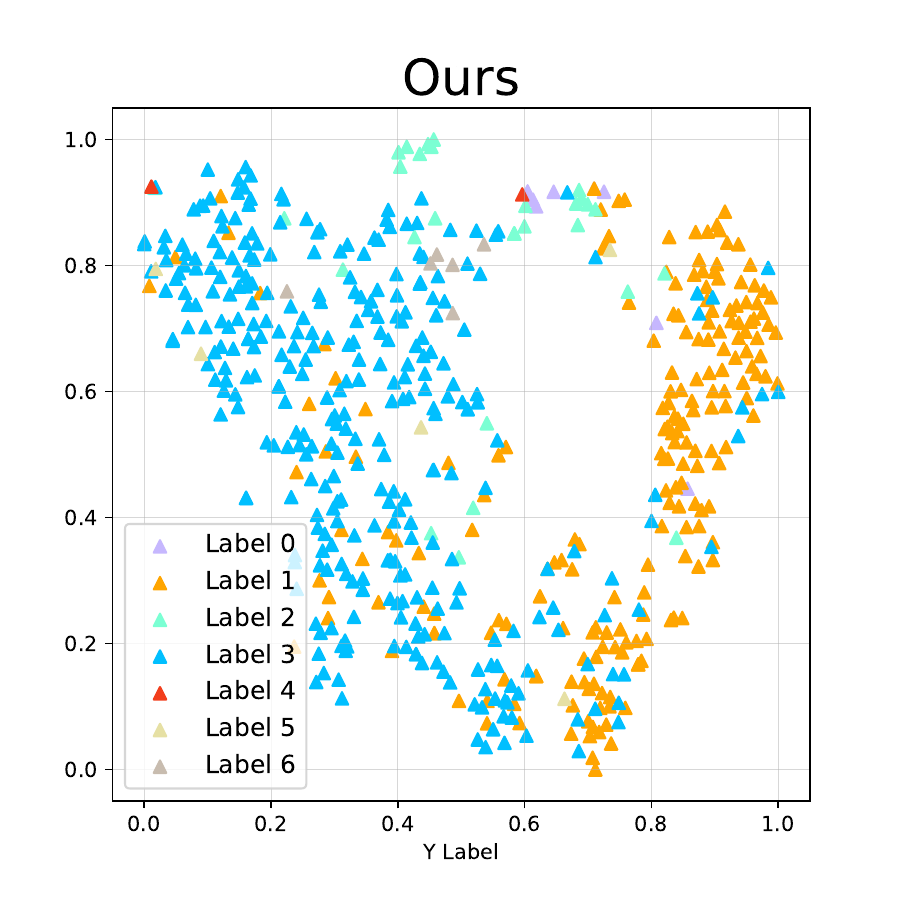}}
\caption{Visualization of feature representation using (a) FLIP (b) $\rm{M^3AE}$, and (b) our UMD on the MELINDA dataset. Our UMD demonstrates a clearer clustering of data, which is beneficial for multi-modal classification.}
\label{fig:tsne-2d}
\end{figure*}

\begin{figure}[t!]
\centering
\subfigure[3D Clustering without TD-Calib]{
\label{3d.1}
\includegraphics[scale=.262]{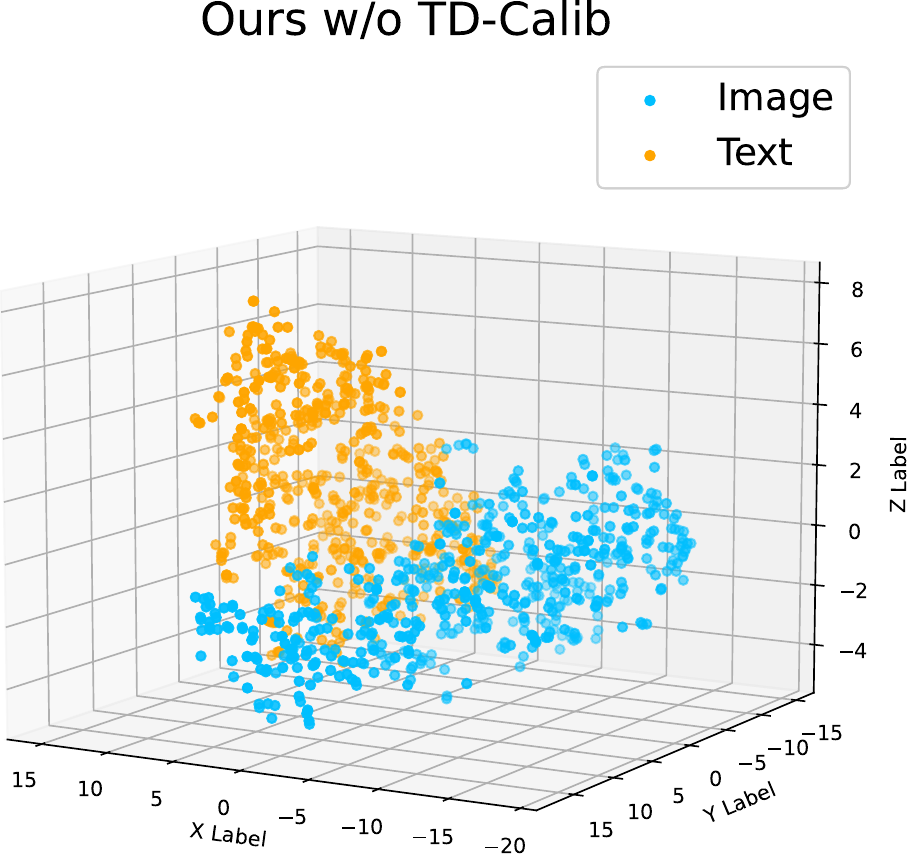}}
\quad
\subfigure[3D Clustering with TD-Calib]{
\label{3d.2}
\includegraphics[scale=.258]{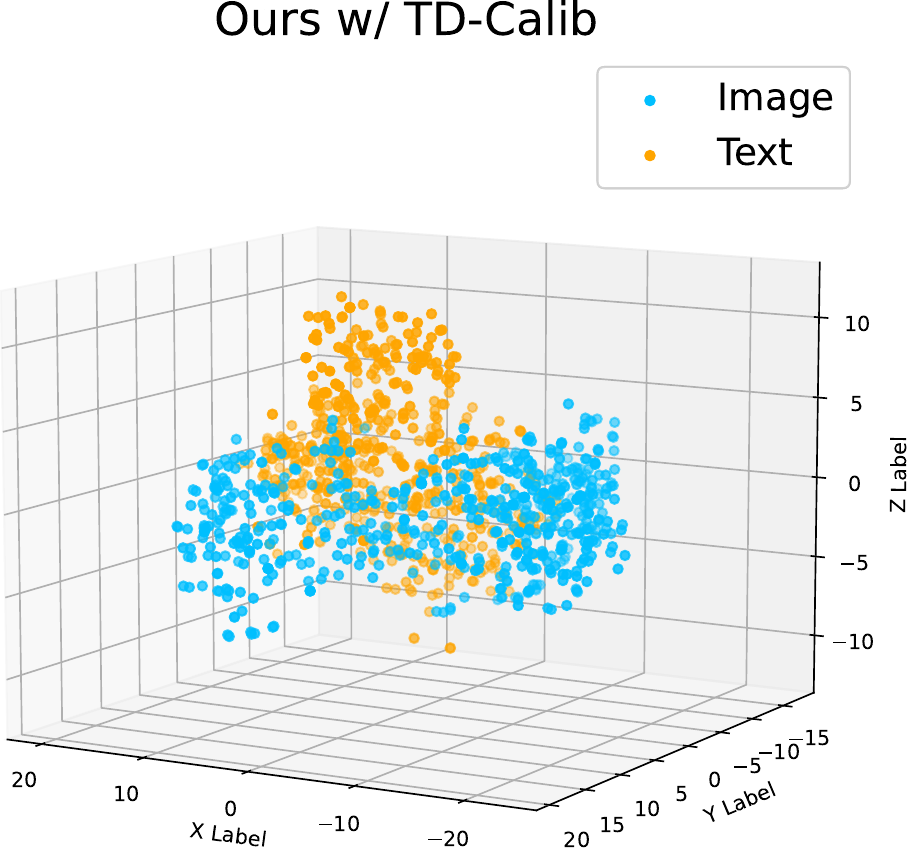}}
\caption{Visualization of image and text embeddings in our UMD framework (a) without or (b) with TD-Calib on MELINDA dataset. The TD-Calib in (b) makes the fusion of multi-modal more adequate.} \label{fig:tsne-3d}
\end{figure}

For the VQA tasks in Table \ref{table:abvqa}, the MR-Pretrain model (Line 2) achieves the \textit{Overall} accuracy of 77.38\%, 83.51\%, and 77.07\% for the VQA-RAD, SLAKE and VQA-Med-2019 datasets, respectively, which exhibits 3.99\% for VQA-RAD, 7.36\% for SLAKE, and 5.07\% for VQA-Med-2019 increase compared with the baseline (Line 1). These improvements can be attributed to the transferable weights learned by the multi-modal encoder model.
Moreover, the \textit{Overall} accuracy increase (Lines 2-4) compared with the baseline (Line 1) in all types of questions on three VQA datasets verifies the effectiveness of the three proposed components (\textit{i.e.}, MR-Pretrain, TD-Calib, GM-Coord). 
Furthermore, the \textit{Overall} performance of the model with two components (Lines 5-7) is better than the results of the model with one component (Lines 2-4), which confirms the complementary enhancement of the proposed components. In addition, the complete UMD framework (Line 8) achieves the best \textit{Overall} performance, validating the effectiveness of our UMD framework.

For the medical image-text classification task, we perform the ablation study of UMD to investigate the combination of pre-training and fine-tuning techniques with various metrics on the MELINDA datasets, as shown in Table \ref{table:abcls}. Similar to the conclusion in Table \ref{table:abvqa}, the results in Lines 2-4 in Table \ref{table:abcls} outperform the baseline (Line 1), which demonstrates each of our designs is rational. Especially, the MR-Pretrain improves by 12.46\% compared with the baseline (Line 1), showing the effectiveness of our pre-training method. Furthermore, compared with models with a single design (Lines 2-4), models of pairwise combination (Lines 5-7) deliver higher performance, which demonstrates the complementarity of the proposed three components. Finally, when all three proposed components are applied (Line 8), our UMD framework achieves the best performance. The ablation study verifies the effectiveness of our MR-Pretrain, TD-Calib and GM-Coord modules.

\subsection{Qualitative Analysis}
For a qualitative comparison, we further present $5$ VQA test samples from the VQA-RAD dataset to provide predicted results of $\rm{M^3AE}$ and our UMD framework, as shown in Fig. \ref{fig:vqa2019test}. Compared with $\rm{M^3AE}$, UMD can understand diagnosis-related information better and predict more accurate answers, which can benefit clinical diagnosis more effectively.

Furthermore, we visualize the t-SNE features \cite{van2008visualizingtSNE} of randomly sampled cases in the MELINDA dataset, as depicted in Fig. \ref{fig:tsne-3d}, where the blue and orange points represent image features $Z^I$ and text features $Z^T$, respectively. By comparing (a) and (b) in Fig. \ref{fig:tsne-3d}, we observe that the fusion of two modalities is more adequate in the case with TD-Calib. This finding highlights another benefit of TD-Calib by enhancing modality fusion, which explains from another perspective why TD-Calib can contribute to various multi-modal downstream tasks.
Additionally, we perform the 2D clustering of FLIP, $\rm{M^3AE}$, and our UMD framework, and visualize the results in Fig. \ref{fig:tsne-2d}. The different colors represent different categories in the MELINDA dataset. The comparison between (a), (b), and (c) in Fig. \ref{fig:tsne-2d} indicates that our UMD framework can separate the categories more distinctly.

\section{Conclusion}
In this work, we propose the Unified Medical Multi-modal Diagnostic (UMD) framework, which utilizes unlabeled multi-modal medical datasets to enhance the representation learning of deep learning models in a self-supervised manner. Specifically, we devise a novel MR-Pretrain strategy, which guides models to capture semantic information from masked inputs of various modalities through feature-level and data-level reconstruction. Moreover, to tackle the distribution heterogeneity between pre-training and downstream data and the modality heterogeneity within downstream datasets, we present a heterogeneity-combat downstream tuning strategy, including the TD-Calib and the GM-Coord. In particular, the TD-Calib fine-tunes the pre-trained model based on the distribution of the downstream datasets, while GM-Coord adjusts the gradient weights according to the dynamic optimization status of different modalities. Extensive experiments on five public medical datasets demonstrate the effectiveness of our UMD framework, which outperforms state-of-the-arts on three kinds of downstream tasks by a remarkable margin.

\appendices
\section*{References}
\bibliographystyle{IEEEtran}
\bibliography{IEEEabrv,refs}

\end{document}